\documentclass{article}

\usepackage[final]{neurips_2021}

\usepackage[utf8]{inputenc} % allow utf-8 input
\usepackage[T1]{fontenc}    % use 8-bit T1 fonts
\usepackage{hyperref}       % hyperlinks
\usepackage{url}            % simple URL typesetting
\usepackage{booktabs}       % professional-quality tables
\usepackage{amsfonts}       % blackboard math symbols
\usepackage{nicefrac}       % compact symbols for 1/2, etc.
\usepackage{microtype}      % microtypography
\usepackage{xcolor}         % colors

\setcitestyle{square,numbers}
\usepackage{url}            % simple URL typesetting
\usepackage{booktabs}       % professional-quality tables
\usepackage{array}
\newcolumntype{L}[1]{>{\centering\arraybackslash}m{#1}}

\usepackage{multirow}
\usepackage{threeparttable}
\usepackage{caption}
\usepackage{subcaption}
\usepackage{graphicx}
\usepackage{float}
\usepackage{amssymb}   % Use this package instead of package {gensymb} to use command \geqslant
\usepackage{amsmath}

\usepackage[ruled, linesnumbered, vlined]{algorithm2e}  % Package option [ruled] makes the algorithm title appear on top with a horizontal line
\SetArgSty{textnormal}
\newcommand{\llIf}[2]{{\let\par\relax\lIf{#1}{#2}}}
\newcommand{\llElse}[1]{{\let\par\relax\lElse{#1}}}
\newcommand{\forcond}{}

\title{Automatic Curricula via Expert Demonstrations}

\author{%
  Siyu Dai, Andreas Hofmann, Brian Williams \\
  Computer Science and Artificial Intelligence Laboratory\\ 
  Massachusetts Institute of Technology, USA\\
  \texttt{sylviad@mit.edu} \\
}

\begin{document}

\maketitle

\begin{abstract}
  We propose Automatic Curricula via Expert Demonstrations (ACED), a reinforcement learning (RL) approach that combines the ideas of imitation learning and curriculum learning in order to solve challenging robotic manipulation tasks with sparse reward functions. Curriculum learning solves complicated RL tasks by introducing a sequence of auxiliary tasks with increasing difficulty, yet how to automatically design effective and generalizable curricula remains a challenging research problem. ACED extracts curricula from a small amount of expert demonstration trajectories by dividing demonstrations into sections and initializing training episodes to states sampled from different sections of demonstrations. Through moving the reset states from the end to the beginning of demonstrations as the learning agent improves its performance, ACED not only learns challenging manipulation tasks with unseen initializations and goals, but also discovers novel solutions that are distinct from the demonstrations. In addition, ACED can be naturally combined with other imitation learning methods to utilize expert demonstrations in a more efficient manner, and we show that a combination of ACED with behavior cloning allows pick-and-place tasks to be learned with as few as 1 demonstration and block stacking tasks to be learned with 20 demonstrations.
\end{abstract}

\section{Introduction}

Imagine a robot factory is trying to manufacture home support robots and sell them all over the world. When the customer asks the robot to complete a novel task, he will likely not be willing to program a detailed task specification or carefully design a set of rewards to guide the robot. What the customer is more likely to provide is probably a high-level goal, a handful of demonstrations, or a combination of both. 
% In order for intelligent robots to be deployed in a variety of real-world applications, it is important that they don't heavily rely on carefully engineered task specifications or intensive human supervision. 
Classical task and motion planning solutions to robotic manipulation~\cite{kaelbling2013integrated} often require carefully engineered domain specifications, but it is infeasible to pre-define all possible tasks the robot might be asked to do in all possible environments. Reinforcement learning (RL) approaches don't require the domain model, though they typically only work well in well-structured environments with carefully designed dense reward signals~\cite{levine2016end}. 
% Imagine a robot factory is trying to manufacture home support robots and sell them all over the world, then it would be infeasible to pre-define all possible tasks the robot might be asked to do or build in a set of reward-shaping functions for each potential scenario. In these cases, it is more preferable to if the customers only need to provide instructions via a binary reward signal and a small number of demonstrations.
Solving RL problems with only sparse or binary rewards has been a long-standing challenge for researchers and many approaches have been proposed, including intrinsic motivation~\cite{pathak2017curiosity, colas2019curious, DaiS-RSS-21}, hierarchical RL~\cite{bacon2017option, haarnoja2018latent} and curriculum learning~\cite{bengio2009curriculum, ivanovic2019barc}. On the other hand, imitation learning~\cite{ho2016generative, peng2018variational} methods resort to expert demonstrations instead of hand-designed reward signals and have shown impressive performance, especially in tasks where the reward functions are tricky to define but demonstrations are easier to obtain. In many tasks in robotics, both a binary reward for tasks success and a small amount of demonstrations can be provided easily, so can we use demonstrations to overcome the challenging exploration problem RL agents face in these long-horizon sparse-reward tasks?

An intuitive idea of utilizing human demonstrations to overcome exploration challenges is to automatically generate a curriculum through demonstration trajectories. With a well-designed curriculum, RL agents can first solve simpler problems where rewards are easy to obtain in order to master the skills that can increase their chance of getting rewards in the challenging tasks. In many manipulation tasks, designing curricula can be tricky and tedious, but human demonstrations can naturally be converted into curricula. Suppose we have a demonstration trajectory $\tau = (s_0, s_1, \ldots, s_{T-1}, s_T)$, where $s_0$ is the initial state and $s_T$ is the goal state. If we assume the demonstration trajectory solves the task in a reasonable way without deliberate detours (even though it could be suboptimal), then it is also reasonable to assume that $s_{T-1}$ is closer to $s_T$ in the task space than $s_1$, which means an RL agent starting from $s_{T-1}$ will likely have a higher chance of reaching $s_T$ than an agent starting from $s_1$ within a limited time of random exploration. Therefore, among all tasks with binary rewards that are only given when the goal $s_T$ is fully reached, the ones with agents initialized at $s_{T-1}$ should be easier than the ones with agents initialized at $s_1$. We hypothesize that agents who have already learned how to reach $s_T$ from $s_{T-1}$ can provide a warm start for agents trying to reach $s_T$ from $s_1$, and that these tasks starting from different initial states can form a systematic curriculum for learning challenging long-horizon tasks with sparse rewards. 

With this intuition, we propose Automatic Curricula via Expert Demonstrations (ACED), a RL approach which uses states from different sections along demonstration trajectories as reset states and controls the curriculum by moving reset states from the end of the demonstrations to the beginning based on the agent's performance. 
Although prior works~\cite{salimans2018learning, resnick2018backplay} evaluated similar ideas in grid world environments and games, the ability of utilizing an arbitrary number of demonstrations and generalizing to random unseen initializations and goals was not provided.
% Although similar ideas have been previously evaluated in grid world environments and games~\cite{salimans2018learning, resnick2018backplay}, no prior work has demonstrated its performance in continuous control tasks and its ability to generalize to random unseen initializations and goals. 
In this paper, we evaluate ACED in robotics pick-and-place tasks and block stacking tasks with only binary rewards, two challenging tasks in the continuous control domain that haven't been solved by vanilla RL algorithms, and analyze the influence of the number of demonstrations and the total number of sections the demonstrations are divided into on ACED's performance. An additional advantage of ACED is that it can be naturally combined with many other methods of utilizing human demonstrations in order to further improve its performance or reduce the number of demonstration trajectories needed, and a combination of ACED and behavior cloning (BC) is demonstrated as an example in this paper. Empirical results show that pick-and-place can be learned with as few as 1 demonstration, and block stacking can be learned with as few as 20 demonstrations. 

\section{Related Work}

\subsection{Curriculum Learning}

Curriculum learning~\cite{bengio2009curriculum} is a continual learning method that accelerates the learning progress by gradually increasing the task difficulty.
It has seen success in many applications including language modeling~\cite{graves2017automated}, autonomous navigation~\cite{mirowski2018learning, ivanovic2019barc} and robotic manipulation~\cite{florensa2017reverse}. 
However, many curriculum-based methods only involve a small and discrete set of manually generated task sequences as the curriculum, and existing automated curriculum generating methods often assume prior knowledge on how to manipulate the environment~\cite{wang2019paired}, or inherit the instability of adversarial methods and bias the exploration to a small subset of the tasks~\cite{florensa2018automatic, sukhbaatar2018intrinsic}. \cite{florensa2017reverse} introduced the idea of automatically generating initial states closer to the goal state in order to speed up training, but inevitably face the challenge of infeasible randomly-generated initial states and can't be trivially extended to problems where the action space distance is not a good indicator of task difficulty. In order to address these issues, we propose to use states from expert demonstrations as initial states to guarantee feasibility and provide more accurate indication of task difficulty.

\subsection{Learning from Demonstration}

Learning from demonstration (LfD) is widely used in tasks where the reward function is hard to define but demonstrations are relatively easier to obtain. Behavior cloning~\cite{pomerleau1991efficient, bain1995framework} is a classical LfD approach that utilizes supervised-learning to train agents that imitate demonstration behaviors. Although BC has seen success in various fields including autonomous driving~\cite{Ogale-RSS-19, wang2019deep} and robotics manipulation~\cite{rahmatizadeh2018vision}, it inevitably demands a large amount of demonstrations and its performance often suffers from data distribution mismatch~\cite{ross2011reduction}. Inverse reinforcement learning~\cite{ng2000algorithms, finn2016guided} infers the reward function through demonstrations in order to avoid manual reward engineering, but it is fundamentally challenging due to its ambiguity in solutions since one trajectory can often be explained by many different reward functions~\cite{fu2017learning}. LfD approaches based on Generative Adversarial Networks (GAN)~\cite{ho2016generative, arjovsky2017wasserstein, peng2018variational} have effectively scaled to applications with relatively high-dimensional environments, but challenges due to unstable GAN training have significantly restricted their success in long-horizon tasks with complicated environments. Another popular approach to effectively utilize expert demonstrations is to combine LfD with RL~\cite{vecerik2017leveraging, kang2018policy, nair2018overcoming}. The advantage of ACED is that it can easily be combined with many existing LfD methods for more efficient utilization of expert demonstrations, including BC, GAN-based methods~\cite{ho2016generative} and methods that add demonstrations in RL replay buffers~\cite{vecerik2017leveraging, nair2018overcoming}. In the empirical evaluation section in this paper, we demonstrate the performance of our approach when combined with BC and show that it provides better convergence performance compared to using ACED only.

\subsection{State Resetting}

State resetting is widely used in RL for introducing expert knowledge, applying curricula or providing safety guarantees. \cite{andrychowicz2017hindsight} and \cite{nair2018overcoming} reset some training episodes to states from expert demonstrations to simplify the exploration challenges in long-horizon tasks. \cite{eysenbach2018leave} and \cite{xu2020continual} show that learning both the forward policy and the reset policy can not only reduce human effort in real-world robotics training but also accelerate training by automatically forming a curriculum. \cite{turchetta2020safe} introduces ``teacher's interventions'' via state resetting to avoid costly mistakes during learning in safety-critical applications. Similar to our proposed approach, \cite{salimans2018learning} and \cite{resnick2018backplay} reset the initial states during training to demonstration states in order to form a sequence of curricula. However, \cite{salimans2018learning} pointed out its limitations in terms of generalizing to unseen states and didn't provide evaluation on continuous control tasks or in-depth analysis on different component's influence on the overall performance, whereas \cite{resnick2018backplay} used a set of fixed rules for switching curricula instead of adapting it based on the agent's performance. In contrast to \cite{salimans2018learning} and \cite{resnick2018backplay} which can only utilize one demonstration trajectory, ACED allows for arbitrary numbers of demonstrations through trajectory sectioning. In this paper, we evaluate ACED on continuous control tasks and analyze the influence of the number of demonstration trajectories and the number of sections they are divided into on ACED's overall performance.

\section{Preliminaries}

\subsection{Reinforcement Learning}

The problem studied in this paper is formulated as a Markov Decision Process (MDP) defined by states $\mathbf{s} \in \mathcal{S}$, actions $\mathbf{a} \in \mathcal{A}$, a transition model $T: \mathcal{S} \times \mathcal{A} \times \mathcal{S} \rightarrow \mathbb{R}$, and a reward function $r: \mathcal{S} \times \mathcal{A} \rightarrow \mathbb{R}$. $\mathcal{S}$ and $\mathcal{A}$ represent the state space and the action space respectively. The objective of the RL problem is to find a policy $\pi: \mathcal{S} \rightarrow \mathcal{A}$ that maximizes $J=\mathbb{E}_{\pi}[\sum_{\tau} r(\mathbf{s}_t, \mathbf{a}_t) | \mathbf{a}_t \sim \pi (\mathbf{s}_t), \mathbf{s}_0 \sim p_0(\mathbf{s})]$, where $\tau$ denotes the rollout trajectory~\cite{sutton2018reinforcement}. ACED can work with any standard RL algorithm, and in this paper we demonstrate its performance using Proximal Policy Optimization (PPO)~\cite{schulman2017proximal} algorithm and Deep Deterministic Policy Gradient (DDPG)~\cite{lillicrap2015continuous}.
% In this paper, we refer to the reward from the environment as extrinsic reward $r^e$ and the artificial reward from the algorithm as intrinsic reward $r^i$, hence $r = r^e + r^i$. The sum $r$ is used during the learning process, whereas only $r^e$ is considered when evaluating the performance of a learning algorithm.

\subsection{Behavior Cloning}

Given expert demonstration trajectories $\mathcal{T} = \{\tau_1, \ldots, \tau_N\}$ where each trajectory includes state-action pairs, i.e. $\tau_e = (\mathbf{s}_0, \mathbf{a}_0, \ldots, \mathbf{s}_t, \mathbf{a}_t, \ldots, \mathbf{s}_T)$, the objective of BC is to learn a mapping from states to actions through supervised learning in order to imitate expert behaviors. Due to BC's demand for a large amount of demonstrations and its poor generalization performance in unseen states, it is often used to pre-train the policy network for other imitation learning or RL approaches as a warm start instead of as a standalone imitation learning approach. The ACED method in this paper can also be combined with BC by using it to pre-train policy networks, and we present this combination in Section~\ref{sec:approach}. We compare the performances of ACED with or without BC in Section~\ref{sec:results}. Note that demonstration trajectories with state-action pairs are only required by BC, and if ACED is used without BC, then demonstration trajectories with only states are sufficient.

\begin{figure}
 \begin{center}
 \includegraphics[width=0.9\linewidth]{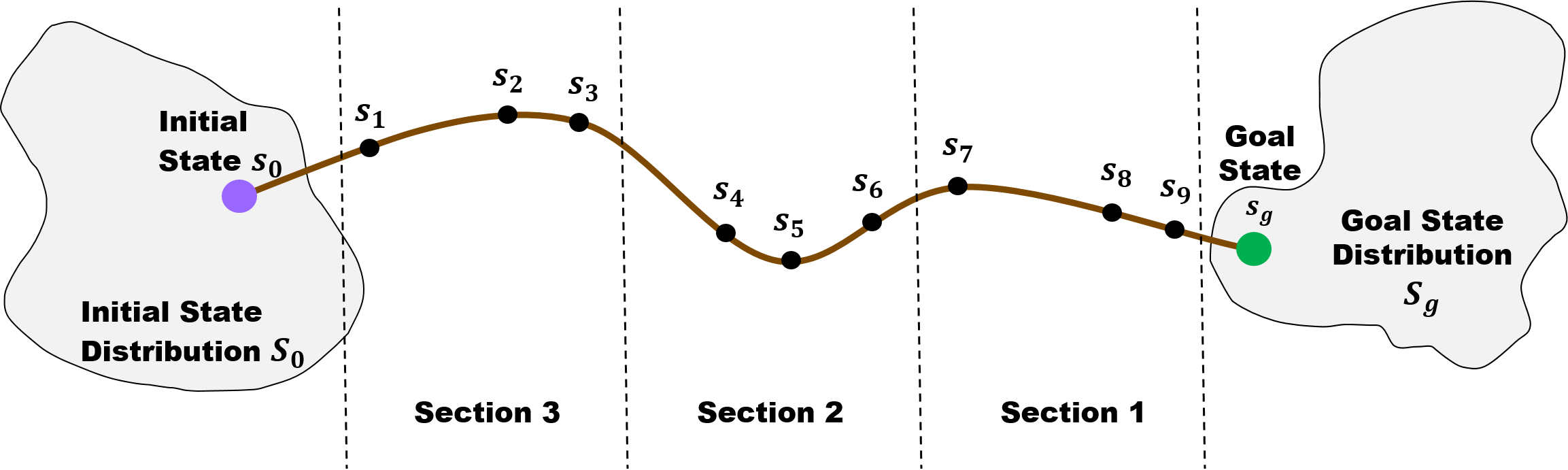}
 \caption{Example of demonstration trajectory segmentation: an expert demonstration trajectory can be divided into sections where larger section number indicates being closer to the initial state. The total number of sections is also called the total number of curricula $C_{max}$, and in this example $C_{max} = 3$. Normal rollout workers randomly sample an initial state from the initial state distribution $S_0$ and a goal state from the goal state distribution $S_g$. For curriculum rollout workers, the environments are reset based on the curriculum number $C$: curriculum-$C$ tasks reset the environment to a section-$C$ state on a randomly selected demonstration trajectory. ACED starts training with $C=1$ and gradually moves reset states towards the beginning of demonstration trajectories by increasing $C$. When ACED switches to normal rollout workers, the reset states are drawn from the actual $S_0$ the target task specifies, and this is when it starts to generalize to unseen initializations.}
 \label{fig:sections}
 \end{center}
 \vspace{-5.5mm}
\end{figure}

\section{Approach: Automatic Curricula via Expert Demonstrations (ACED)} \label{sec:approach}

In order to solve long-horizon manipulation tasks with binary rewards, ACED constructs a curriculum by sampling states from expert demonstration trajectories as initializations for each training episode, where the samples initially come from near the end of the demonstration trajectories and gradually move forward as the agent improves its performance.

\begin{center}
\begin{minipage}{.8\linewidth}
 \begin{algorithm}[H]
 \caption{Automatic Curricula via Expert Demonstrations}
 \label{algorithm:full}
 \small
 \DontPrintSemicolon
 \KwIn{\\ $T$: \text{number of iterations}
 \\ $C_{max}$: \text{total number of curricula}
 \\ $\mathcal{T} = \{\tau_1, \ldots, \tau_N\}$: \text{demonstration trajectories}
 \\ $\phi$: \text{curriculum switching threshold for average return}
 \\ $t$: \text{period for checking average return}
 \\ $n$: \text{number of episodes used to compute average return}
}
 \KwOut{%\\ $c$: \text{collision number}
 \\ $\pi$: \text{policy} }
 
 Initialize policy parameters with Behavior Cloning Algorithm \\
 Initialize curriculum number $C \leftarrow 1$ \\
 Initialize rollout worker $W \leftarrow$ CurriculumRolloutWorker\\
 Initialize rollout trajectory buffer $\mathcal{E} \leftarrow \{\}$ \\
 \For{\forcond $i=1, 2, \ldots, T$}{
%  Collect experience with a number of parallel Rollout Workers\\
%  $\tau \leftarrow W.\text{rollout}(C, C_{max}, \mathcal{T}, \lambda)$ \\
 $\tau \leftarrow W.\text{rollout}(C, C_{max}, \mathcal{T}, \pi)$ \\
 Add $\tau$ to $\mathcal{E}$ \\
 Send $\mathcal{E}$ to RL Algorithm and update $\pi$ \\
 \If{$i \mod t == 0$}{
 $R \leftarrow$ Evaluate the average return on the most recent $n$ episodes\\
 \If{$R \geqslant \phi$}{
 \eIf{$C < C_{max}$}{$C \leftarrow C + 1$}{
 $W \leftarrow$ NormalRolloutWorker}}
 % Check whether to update curriculum\\
 }}
\end{algorithm}
\end{minipage}
\end{center}

We first collect a set of expert demonstration trajectories $\mathcal{T}$ and represent each trajectory $\tau_e \in \mathcal{T}$ as a discrete sequence of states at each time step: $\tau_e = (\mathbf{s}_0, \mathbf{s}_1, \ldots, \mathbf{s}_{T-1}, \mathbf{s}_T)$, where the initial state $\mathbf{s}_0$ is randomly sampled from the initial state distribution $S_0$ and the final state $\mathbf{s}_T$ has a probability of $p_{success}$ to reach a goal state randomly sampled from the goal distribution $S_g$, i.e. $\mathbb{P} (\mathbf{s}_T = \mathbf{s}_g) = p_{success}, \mathbf{s}_g \in S_g$. We refer to $p_{success}$ as the expert success rate.
Each trajectory $\tau_e$ is then evenly divided into $C_{max}$ sections, where section-$C_{max}$ denotes the section at the beginning of $\tau_e$ (near $\mathbf{s}_0$) and section-$1$ denotes the one at the end (near $\mathbf{s}_T$). $C_{max}$ is a hyperparameter referred to as the \emph{total number of curricula}. Figure~\ref{fig:sections} provides an illustration of an example demonstration trajectory and its segmentation. A key assumption made in ACED is that all expert demonstrations are reasonable solutions to the problem and don't contain unnecessary detours, which guarantees that two states that are close in the demonstration trajectory are also close in the task space. 

\begin{center}
\begin{minipage}{.8\linewidth}
\begin{algorithm}[H]
 \caption{CurriculumRolloutWorker}
 \label{algorithm:env}
 \small
 \DontPrintSemicolon
 \KwIn{\\ $C$: \text{current curriculum number}
 \\ $C_{max}$: \text{total number of curricula}
 \\ $\mathcal{T} = \{\tau_1, \ldots, \tau_N\}$: \text{demonstration trajectories}
%  \\ $\lambda$: \text{curriculum switching threshold for average return}
 \\ $\pi$: \text{current policy}
}
 \KwOut{%\\ $c$: \text{collision number}
 \\ $\tau$: \text{rollout trajectory} }
 Randomly select a trajectory from demonstrations $\tau_e \in \mathcal{T}$ \\
 $num\_transitions = \text{len}(\tau_e) - 1$ \tcc*[r]{Make sure to not sample the goal state}
 $interval = \text{RoundDown}(num\_transitions / C_{max})$ \tcc*[r]{Divide $\tau_e$ into $C_{max}$ intervals}
 $index = \text{RandInt}(interval) + interval \times (C_{max} - C)$ \tcc*[r]{Randomly select a state from the segment of $\tau_e$ corresponding to the current $C$}
 $\mathbf{s}_{init} = \tau_e[index]$  \\
 $\tau = \text{Rollout}(env, \mathbf{s}_{init}, \pi)$ \\
\end{algorithm}
\end{minipage}
\end{center}

Algorithm~\ref{algorithm:full} describes the overall framework of ACED. Given the expert demonstration set $\mathcal{T} = \{\tau_1, \ldots, \tau_N\}$, we can pretrain the policy network with BC (line 1). We refer to this version of the algorithm as ACED with BC, and if Algorithm~\ref{algorithm:full} is executed without line 1, then we call it ACED without BC. The performances of the two versions are compared in pick-and-place tasks in Section~\ref{sec:picknplace}.
We use a set of parallel environments to generate rollout data for RL training. We refer to the original environment that resets to initial states sampled from $S_0$ as the \emph{normal rollout worker}, and the curriculum-based environment that resets to demonstration states the \emph{curriculum rollout worker}. Details of the curriculum rollout worker is described in Algorithm~\ref{algorithm:env}. At the beginning of training, all parallel environments are set to be curriculum rollout workers (as shown in Algorithm~\ref{algorithm:full} line 3), and the switch from curriculum rollout workers to normal rollout workers are controlled by \emph{curriculum number} $C \in \{1, 2, \ldots, C_{max}\}$. At each iteration, Algorithm~\ref{algorithm:full} will collect training data using rollout workers and optimize the policy using the RL algorithm of choice (line 6 - 8). $C$ is initialized to be 1, and every $t$ iterations the algorithm will check the average return from the most recent $n$ episodes and compare it with a threshold $\phi$ (line 9 - 15). When the average return exceeds the threshold $\phi$, we add 1 to the current curriculum number $C$ if it hasn't reached $C_{max}$, or switch to the normal rollout worker if $C$ has reached $C_{max}$.

At each rollout, as shown in Algorithm~\ref{algorithm:env}, the curriculum rollout worker will first randomly select a demonstration trajectory $\tau_e$ and divide it into $C_{max}$ sections with equal number of states (line 3). Based on the current curriculum number $C$, the curriculum rollout worker resets the environment to a randomly selected state from section-$C$ on the demonstration trajectory $\tau_e$ (line 4 and 5). It will then rollout a trajectory using the current policy and return it to Algorithm~\ref{algorithm:full}.
% As training goes on, the agent will constantly inspect the learning progress and move the reset state to the previous section if the average return hits a threshold. After the agent successfully learned to complete the pick-and-place tasks from section $C_{max}$ reset states, the environment will be switched to the normal rollout worker instead of the curriculum-rollout worker. 
In our implementation, each rollout worker keeps track of its own curriculum number and the switch from a curriculum rollout worker to a normal rollout worker is independent from other parallel workers' $C$ value. 
% We found this independent curriculum number implementation helpful for preventing unstable training due to sudden performance drop during curriculum switch.
% Unlike \cite{salimans2018learning}, here the goal is to train a policy that can generalize to random start and goal poses on the desktop, hence resetting the training episodes from states along one single demonstration trajectory will not suffice. Therefore, the influence of the number of expert demonstrations on the performance of this method is one aspect to study in this projects. Another aspect to investigate is how the number of stages $C_{max}$ influence its performance.

\section{Empirical Evaluation}  \label{sec:results}

We evaluate our approach on two tasks in the Fetch environment in OpenAI Gym~\cite{1606.01540}: a pick-and-place task and a block stacking task. The pick-and-place task is adapted from Gym directly and the block stacking task is adapted from~\cite{lanier2019curiosity}. The goal of the pick-and-place task is to move a block randomly placed on the tabletop to a goal pose that could be either in the air or on the tabletop, and the goal of the stacking task is to move two randomly placed blocks to their corresponding goal pose where the yellow block is stacked on top of the green block on the tabletop. The majority of results presented in this section use PPO as the RL algorithm, and we demonstrate ACED's off-policy performance with DDPG in the pick-and-place task. The demonstration trajectories are generated from a hand-coded straight-line policy, i.e. the robot is instructed to follow a straight-line trajectory to reach the object, grasp the object and then follow a straight-line trajectory to reach the goal. For both tasks, we use a binary reward function, where $r=1$ indicates all blocks are within the distance threshold to their corresponding goal poses and $r=0$ otherwise. The episode is terminated immediately if $r=1$ is obtained even if the maximum episode length hasn't been reached. Additional implementation details are presented in Appendix~\ref{sec:details}. In the results presented in this section, we mainly focus on the convergence performance and the success rate, but selected examples of the learning curves in the pick-and-place environment are shown in Appendix~\ref{sec:learning_curves}.

\subsection{Pick-and-Place Tasks} \label{sec:picknplace}

\begin{figure}
 \begin{center}
 \includegraphics[width=\linewidth]{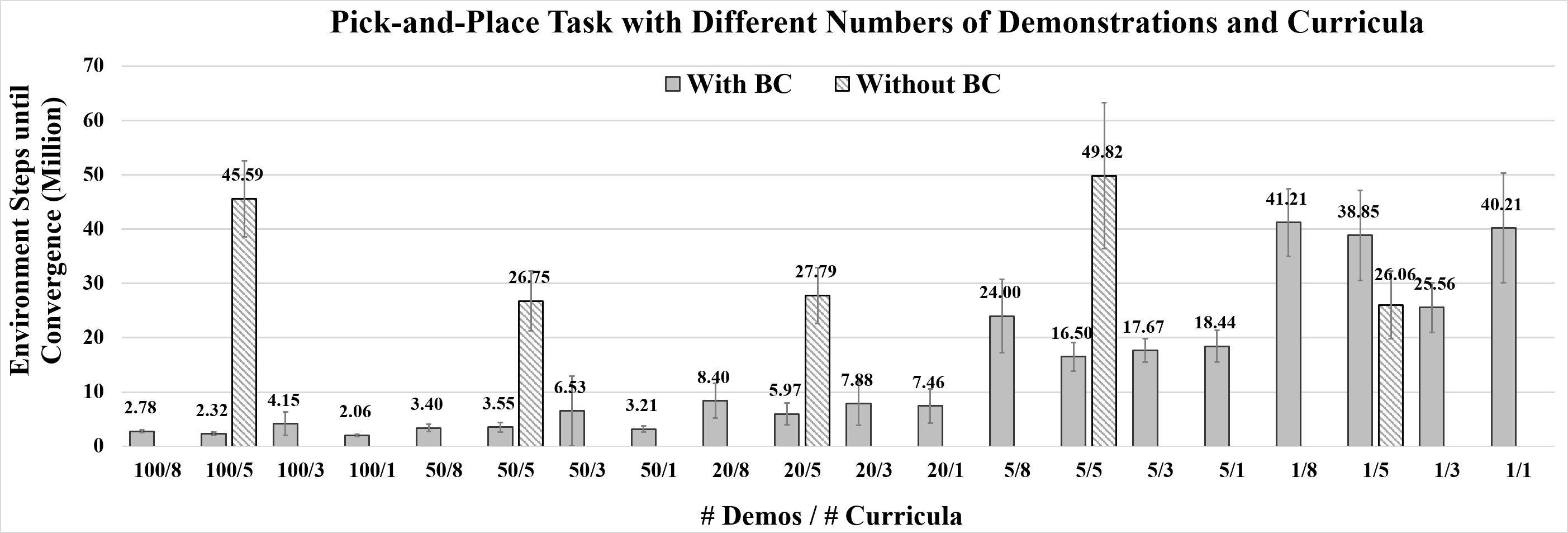}
 \caption{Number of environment steps ACED with BC and ACED without BC take to train pick-and-place tasks with PPO until convergence with different values of the number of demonstration trajectories $|\mathcal{T}|$ and the total number of curricula $C_{max}$. The bars represent the mean of 10 runs with different random seeds and the error bars represent the 90\% confidence interval.}
 \label{fig:BarChart}
 \end{center}
\end{figure}

In this section, we compare the performance of ACED with BC and ACED without BC on pick-and-place tasks in the Fetch environment. In order to study the influence of the number of demonstration trajectories $|\mathcal{T}|$ and the total number of curricula $C_{max}$, we test ACED with BC with $|\mathcal{T}| = 100, 50, 20, 5, 1$ and $C_{max} = 8, 5, 3, 1$. ACED without BC is only tested with $|\mathcal{T}| = 100, 50, 20, 5, 1$ and $C_{max} = 5$, since $C_{max} = 5$ is the best performer in ACED with BC experiments in terms of convergence speed. Here we define convergence as having a training success rate of stably above 90\% and being able to accomplish most tasks during test time. We compare their convergence performance during training with PPO in Figure~\ref{fig:BarChart} and their success rate performance during testing in Table~\ref{table:SuccessRate}. We also tested ACED with DDPG with $|\mathcal{T}| = 5$ and $C_{max} = 5$, and the average steps to convergence is 5.07 million and the success rate is 100\%, proving that ACED can be applied to off-policy RL algorithms and achieve higher sample efficiency.
% We compare the performance in terms of the number of environment steps the training takes to converge in Figure~\ref{fig:BarChart}, and the success rate during testing in Table~\ref{table:SuccessRate}. 
% Due to limited time, the experiments for $|\mathcal{T}| = 5, C_{max} = 3$ and $|\mathcal{T}| = 1, C_{max} = 5, 3$ have not finished, hence are not presented here.

The horizontal axis in Figure~\ref{fig:BarChart} represents the number of demonstration trajectories $|\mathcal{T}|$ and the total number of curricula $C_{max}$, and the vertical axis represents the number of environment steps the training takes to converge averaged from 10 runs with different random seeds. From Figure~\ref{fig:BarChart} we can see that for ACED with BC, $|\mathcal{T}|$ has a significant impact on the number of environment steps it takes to converge. One potential reason that can cause this is, with a more diverse set of initializations during the curriculum training phase, ACED will better generalize to unseen random initial states and goal states after switching to normal rollout worker. Another potential reason is that BC is usually less prone to overfitting when the number of demonstrations is large, hence it should be able to generate better initial policies during pre-training with a larger $|\mathcal{T}|$. In comparison, the number of total curricula has less impact on the convergence performance of ACED with BC, but choosing a reasonable $C_{max}$ can help accelerate training especially when the number of demonstrations is small. For the pick-and-place task we tested on, $C_{max}=5$ generally performs better across different $|\mathcal{T}|$ values in terms of both the mean and the 90\% confidence interval.

\begin{table}
\caption{Pick-and-Place Success Rate with PPO}
\label{table:SuccessRate}
\centering
\begin{threeparttable}
\begin{tabular}{L{2cm}|L{2cm}|L{1.6cm}|L{1.6cm}|L{1.6cm}|L{1.6cm}|L{1.6cm}}
\toprule

\multirow{2}{2cm}{\centering Algorithm} & \multirow{2}{2cm}{\centering Number of Curricula\tnote{1}} 
& \multicolumn{5}{c}{Number of Demonstrations} \\ \cline{3-7}

& & $|\mathcal{T}|=100$  & $|\mathcal{T}|=50$  & $|\mathcal{T}|=20$  & $|\mathcal{T}|=5$  & $|\mathcal{T}|=1$   \\ 

\hline

\multirow{5}{2cm}{\centering ACED with BC} & $C_{max}=8$ & 99\% & 100\%  & 99\%  & 97\% & 96\% \\
& $C_{max}=5$ & 96\% & 99\%  & 99\%  & 100\% & 95\% \\
& $C_{max}=3$ & 100\% & 99\%  & 100\%  & 100\% & 99\% \\
& $C_{max}=1$ & 100\% & 99\%  & 100\%  & 98\% & 99\% \\

\cline{2-7}

& Average\tnote{2} & 98.8\% & 99.3\%  & 99.5\%  & 98.8\% & 97.3\% \\ \hline
ACED without BC & $C_{max}=5$ & 100\% & 95\%  & 100\%  & 93\% & 97\% \\
\hline
\multicolumn{2}{c|}{BC Policy\tnote{3}} & 60\% & 54\% & 24\% & 2\% & 4\% \\
\multicolumn{2}{c|}{Expert Demonstrations} & 92\% & 98\%  & 95\%  & 80\% & 100\% \\

\bottomrule
\end{tabular}
\begin{tablenotes}
\footnotesize
 \item[1] For each set of experiment, we have 10 runs with different random seeds. For each run, we rollout 10 trajectories with the policy at convergence and compute the success rate, hence each entry is computed from a total of 100 rollout trajectories. 
 \item[2] The average success rate for $C_{max}=8$, $C_{max}=5$, $C_{max}=3$ and $C_{max}=1$.
 \item[3] The success rate of the initial policy pre-trained by BC evaluated on 100 rollout trajectories.
\end{tablenotes}
\end{threeparttable}
\end{table}

If we compare the convergence performance of ACED with BC and ACED without BC in Figure~\ref{fig:BarChart}, we can see that ACED without BC generally takes longer to converge except for when there is only 1 demonstration trajectory. This shows that BC pre-training can provide a good initial policy and accelerate ACED training when sufficient demonstration trajectories are provided. However, when $|\mathcal{T}|$ is too small, BC pre-training might adversely affect ACED's performance. Another observation from Figure~\ref{fig:BarChart} is that the convergence performance of ACED without BC does not show a clear trend as $|\mathcal{T}|$ decreases, which means that the increasing trend we see in ACED with BC experiments is more likely to have been caused more by BC rather than ACED itself. One explanation for this observation is that, despite ACED's better generalization performance when $|\mathcal{T}|$ is large, the curriculum training phase itself can become more challenging and takes longer to converge with a larger $|\mathcal{T}|$. This is because ACED faces a more diverse set of initializations when there are more demonstration trajectories. Our experiments show that without BC pre-training, ACED actually performs the best with only 1 demonstration in pick-and-place tasks.

Table~\ref{table:SuccessRate} compares the success rate of ACED during test time with the success rate of expert demonstrations and behavior cloning. We can see that even though the expert demonstrations aren't perfect (i.e. mostly have a success rate of less than 100\%), ACED is able to learn the pick-and-place task with better-than-expert performance. This is because our approach doesn't rely on the expert policy except for the BC pre-training, and it instead tries to come up with its own policy that reaches the goal from states along demonstrations. Therefore, even with suboptimal demonstrations, ACED can still achieve better-than-expert performance. On the other hand, with policies trained only by BC, the success rate is much lower especially when the number of demonstrations is small, proving that our approach utilizes expert demonstrations in a more effective way than BC does. Another observation from Table~\ref{table:SuccessRate} is that ACED generally achieves higher success rate with more demonstrations, but it is notable that even with 1 demonstration trajectory, it can still achieve a success rate of 96\%. This is very encouraging because unlike many other imitation learning approaches that require a large number of demonstrations to work effectively, ACED can succeed with as few as 1 demonstration.

\subsection{Block Stacking Tasks}

% We run block stacking experiments with $|\mathcal{T}| = 100, 20$, and $C_{max} = 12, 8$.

\begin{figure}
      \centering
      \begin{subfigure}[t]{0.4\textwidth}
      \centering
      \includegraphics[width=0.45\linewidth]{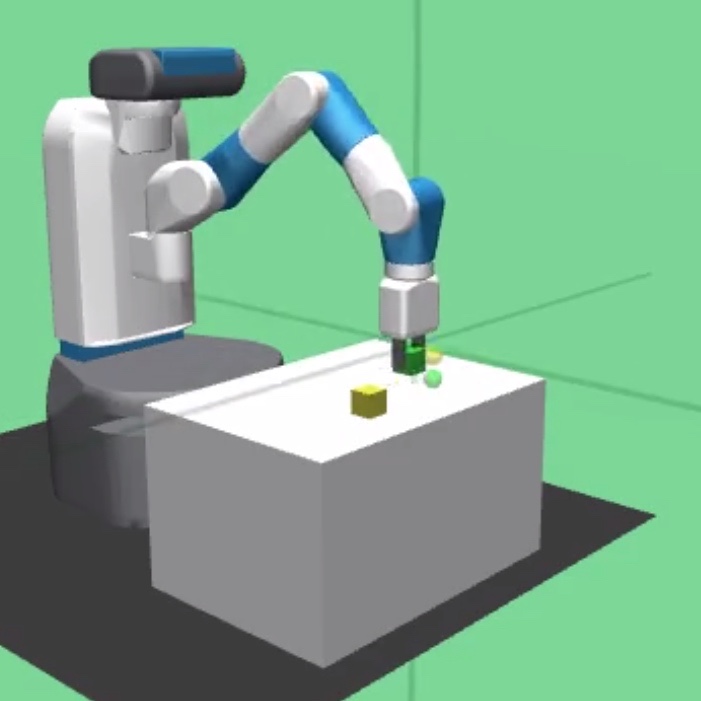}
      \includegraphics[width=0.45\linewidth]{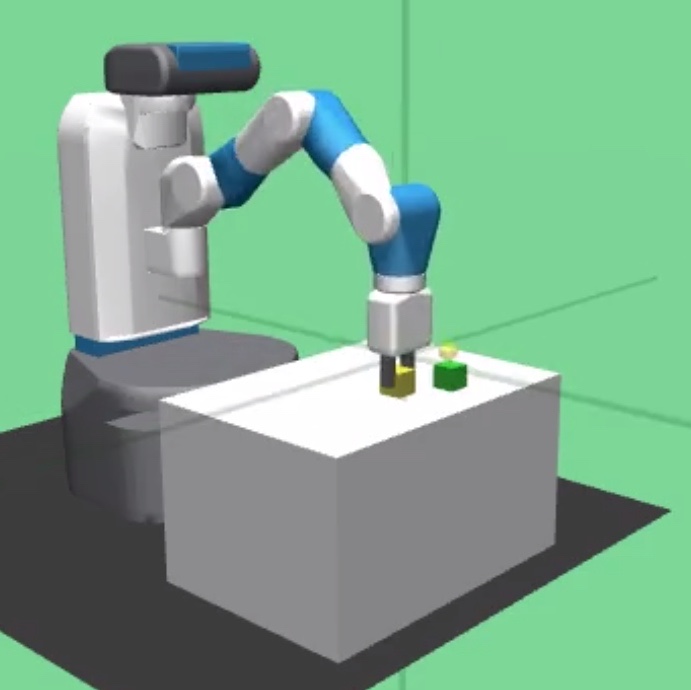}
      \caption{Solution policy when $|\mathcal{T}|=100$}
      \end{subfigure}%
      \begin{subfigure}[t]{0.4\textwidth}
      \centering
      \includegraphics[width=0.45\linewidth]{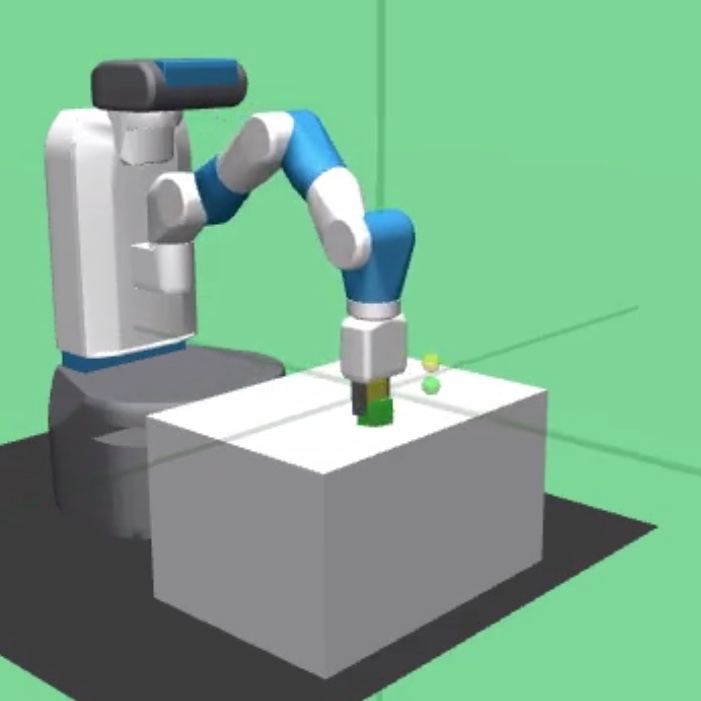}
      \includegraphics[width=0.45\linewidth]{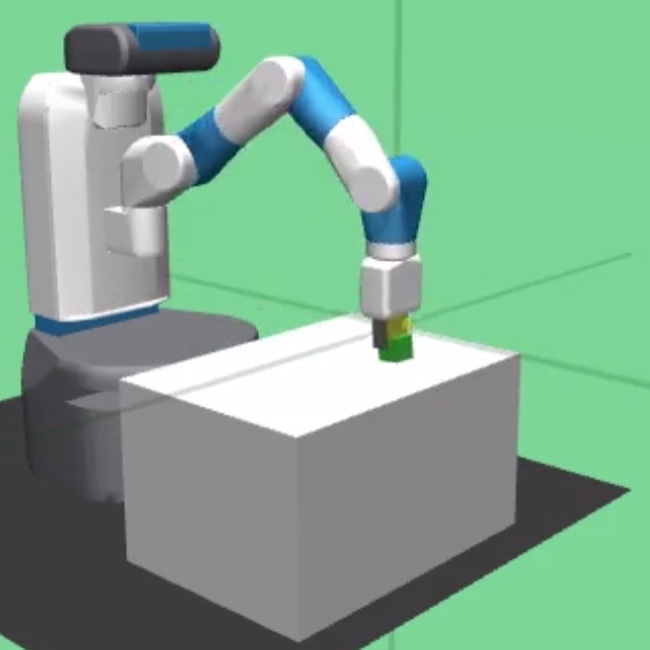}
      \caption{Solution policy when $|\mathcal{T}| = 20$}
      \end{subfigure}%
      \caption{Two different stacking policies ACED with BC converged to with different $|\mathcal{T}|$.}
      \label{fig:stacking}
   \end{figure}

% Due to limited time, we only provide results for two sets of experiments in the block stacking environment: $|\mathcal{T}| = 100, C_{max} = 12$ and $|\mathcal{T}| = 20, C_{max} = 8$.
ACED with BC is also evaluated  in block stacking tasks with $|\mathcal{T}| = 100, 20$ and $C_{max} = 12, 8$, and its performance is compared with other state-of-the-art automatic curriculum methods including reverse curriculum~\cite{florensa2017reverse} and the Montezuma's Revenge method~\cite{salimans2018learning}. Unfortunately, neither of the baseline approaches are able to successfully learn the stacking task (i.e. converge), hence we cannot compare with their convergence performance. 
% We provide more details on their performance and compare the learning curves in Appendix~\ref{sec:learning_curves}. 
Table~\ref{table:Stacking} presents the average number of environment steps ACED with BC takes to converge in each set of experiments, and compares its success rate with the initial policies pre-trained by BC and with the expert demonstrations. Interestingly, we observe that the number of environment steps until convergence is much lower when $|\mathcal{T}| = 20$ compared to when $|\mathcal{T}| = 100$. We believe this is because ACED with BC has converged to two different policies for the two sets of experiments with different $|\mathcal{T}|$ values. From the recorded videos we found that in all 10 runs with $|\mathcal{T}| = 100$ (including $C_{max} = 12$ and $C_{max} = 8$), the policies ACED with BC converged to are similar to the demonstrations, where the robot first picks up the green block and places it onto the goal, and then picks up the yellow block and places it onto the green block. However, in all 10 runs with $|\mathcal{T}| = 20$, the policy ACED with BC converged to takes a different route: it first places the yellow block on top of the green block, and then picks up the two blocks together to place them onto the goal. Figure~\ref{fig:stacking} illustrates the two different policies visually by showing two representative frames from each rollout video. This finding shows that, with a smaller number of demonstrations, ACED with BC has more flexibility to come up with novel solutions instead of following the demonstration trajectories. Because the $|\mathcal{T}| = 20$ solution takes fewer steps, it is easier to train and is more robust when generalizing to new initial and goal poses. We also observed that the $|\mathcal{T}| = 20$ training curves experience less performance drop when switching from curriculum rollout workers to normal rollout workers. We believe this is why ACED with BC converges faster when trained with 20 demonstrations.
% When taking a closer look at the policies, we also noticed that with different $|\mathcal{T}|$ values, ACED with BC converged to different policies. One possible reason that could have caused this is that, with a smaller number of total curricula and a smaller number of demonstrations, the RL agent has more flexibility to come up with novel solutions instead of following the demonstration trajectories. We found that in all 5 runs with $|\mathcal{T}| = 100, C_{max} = 12$, the policy our approach converged to is similar to the demonstrations, where the robot first picks up the green block and places it onto the goal, and then picks up the yellow block and places it onto the green block. However, in all 5 runs with $|\mathcal{T}| = 20, C_{max} = 8$, the policy our approach converged to takes a different route: it first places the yellow block on top of the green block, and then picks up the two blocks together to place them onto the goal. Figure~\ref{fig:stacking} illustrates each of the two different policies visually by showing two representative frames from the rollout videos. 
% Further experiments with different $|\mathcal{T}|$ and $C_{max}$ values are still needed to determine which of the two parameters have more impact on what policy the RL agent converges to.

\begin{table}
\caption{Block Stacking Performance with PPO}
\label{table:Stacking}
\centering
\begin{threeparttable}
\begin{tabular}{L{2cm}|L{3.8cm}|L{1.68cm}|L{1.68cm}|L{1.68cm}|L{1.68cm}}
\toprule
\multicolumn{2}{c|}{\centering Number of Demonstrations}  & \multicolumn{2}{c|}{\centering $|\mathcal{T}|=100$}  & \multicolumn{2}{c}{\centering $|\mathcal{T}|=20$}   \\ 
\hline
\multirow{3}{2cm}{\centering ACED with BC\tnote{1}} & Total Curriculum Number  & $C_{max} = 12$ & $C_{max} = 8$ & $C_{max} = 12$  & $C_{max} = 8$   \\ 
\cline{2-6}

 & Convergence Env Steps\tnote{2} & 213.08 & 169.88 & 143.02 & 119.79 \\
& Success Rate\tnote{3} & 100\% & 100\% & 96\% & 100\% \\
\hline
\multicolumn{2}{c|}{\centering BC Policy Success Rate\tnote{4}} & \multicolumn{2}{c|}{0\%} & \multicolumn{2}{c}{0\%}  \\
\multicolumn{2}{c|}{Expert Demonstration Success Rate\tnote{5}} & \multicolumn{2}{c|}{84\%} & \multicolumn{2}{c}{85\%} \\

\bottomrule
\end{tabular}
\begin{tablenotes}
\footnotesize
 \item[1] Each set of experiment for ACED with BC is averaged from 5 runs with different random seeds. 
 \item[2] Presented in millions. The training for stacking tasks is more unstable, so we separated the training process into two sections: 1) use curriculum rollout workers to train until all parallel workers reach $C=C_{max}$ and 2) set all parallel workers to be normal rollout workers and train until convergence. The convergence environment steps presented here are the sum of the two sections.
 \item[3] The success rate for ACED with BC is evaluated with 10 rollout trajectories per random seed (50 total).
 \item[4] The success rate of the initial policy pre-trained by BC evaluated on 100 rollout trajectories.
 \item[5] The demonstrations for stacking have much lower success rates because there are two blocks in the scene and one block might obstruct the straight line policy the moves the other block to its goal pose. Since the straight line policies are open-loop, the agent can't recover from such failures.
\end{tablenotes}
\end{threeparttable}
\end{table}

Another finding from Table~\ref{table:Stacking} is that for the block stacking task, $C_{max} = 8$ has better performance than $C_{max} = 12$ in terms of both convergence speed and success rate. Compared to the number of demonstrations, the number of total curricula has less impact on ACED with BC's solutions and training performance, and different $C_{max}$ didn't cause the solution policies to differ qualitatively.

\vspace{-2pt}

\begin{figure}
 \begin{center}
 \includegraphics[width=0.8\linewidth]{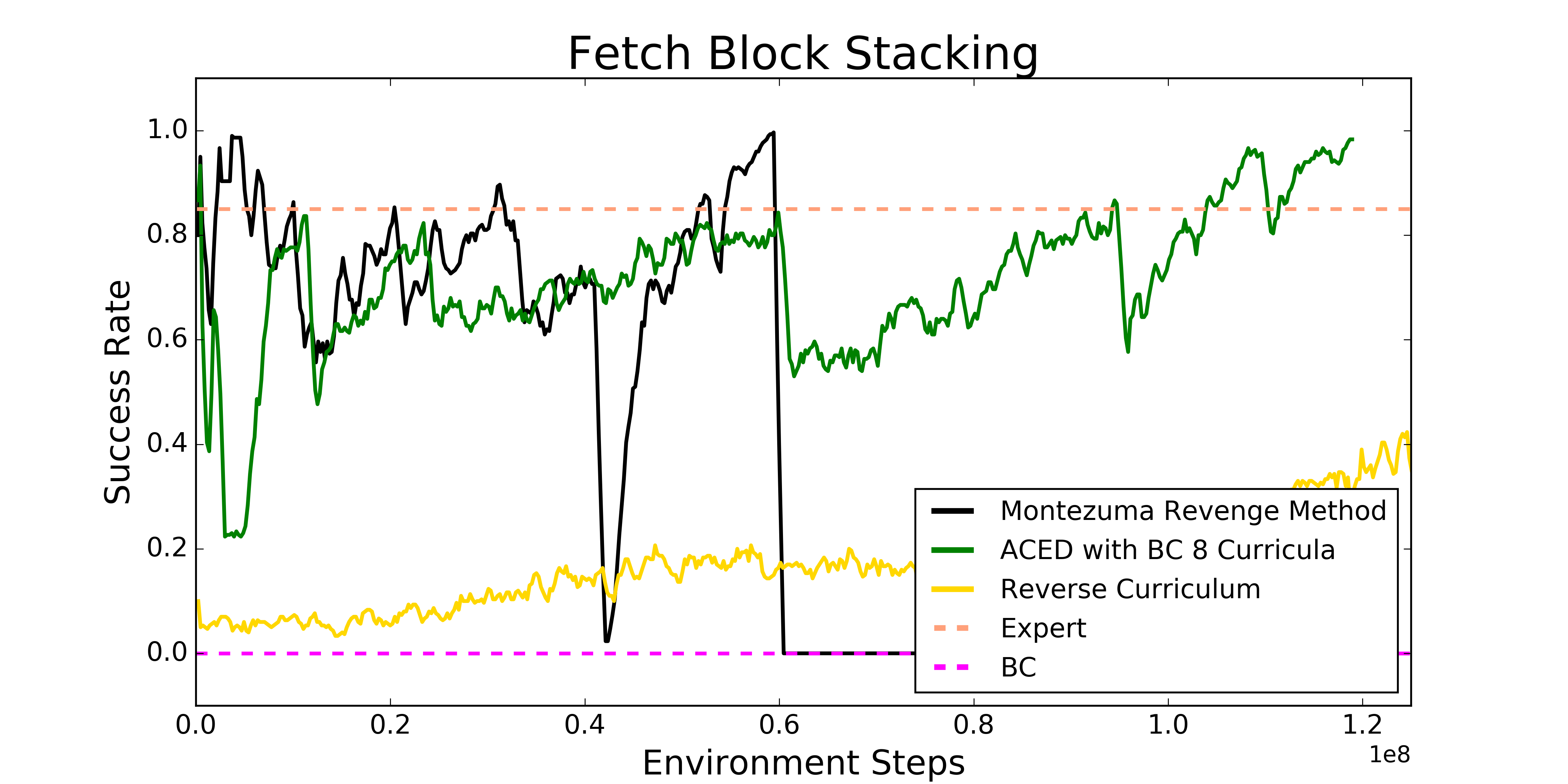}
 \caption{Learning curves of ACED with BC, the reverse curriculum method~\cite{florensa2017reverse}, and the Montezuma's Revenge method~\cite{salimans2018learning} in the block stacking task.}
 \label{fig:stack_curves}
 \end{center}
\end{figure}

We show the comparison of ACED with two state-of-the-art automatic curriculum generation methods~\cite{florensa2017reverse, salimans2018learning} in the block stacking task in Figure~\ref{fig:stack_curves}. We refer to the approach presented in~\cite{florensa2017reverse} the reverse curriculum method, and the one presented in~\cite{salimans2018learning} the Montezuma's Revenge method. In Figure~\ref{fig:stack_curves}, ACED is implemented with PPO and is provided with 20 demonstrations and BC pre-training. 
As we mentioned earlier, neither of the two baseline automatic curriculum methods are able to converge in the block stacking tasks. In all runs of the reverse curriculum method, the start state distribution has not moved to the true $S_0$ after 400 million environment steps of training, and achieves 0\% success rate during testing. In all runs of the Montezuma's Revenge method, moving through the curriculum rollout workers are relatively quick, but the learning curve drops to zero and never goes back up once it switches to the normal rollout worker. This shows that in challenging tasks in continuous state space, policies training using a fixed demonstration without randomization struggle to generalize to the entire initial state distribution $S_0$ and goal distribution $S_g$. Figure~\ref{fig:stack_curves} shows the learning curves of one representative run for each of the curriculum generation methods.

\section{Discussion}

This paper presents Automatic Curricula via Expert Demonstrations, an RL approach that combines ideas from both imitation learning and curriculum learning to tackle challenging robotics manipulation tasks with sparse reward signals. Through resetting the training episodes to states along demonstration trajectories, ACED is able to control the difficulty of the tasks by moving the reset states from the end of the demonstration to the beginning based on the learning progress of the RL agent. This procedure naturally forms a curriculum and makes challenging exploration problems feasible to learn. One main advantage of ACED is that it only requires demonstration states and not actions when deployed on its own. ACED can also be intuitively combined with many existing imitation learning approaches to utilize expert demonstrations more efficiently, including adding demonstrations to replay buffers~\cite{nair2018overcoming}, introducing GAN-based rewards\cite{ho2016generative}, and using behavior cloning to pre-train the policies. In this paper, a version of ACED with policies pre-trained via behavior cloning is compared with ACED on its own as an example. 
We evaluate the performance of ACED on block pick-and-place tasks and stacking tasks, and show that pick-and-place can be learned with as few as 1 demonstration and stacking can be learned with 20 demonstrations. We also analyzed the impact of the number of demonstration trajectories and the total number of curricula on ACED's performance, and discovered that ACED can learn novel solutions that are very distinct from expert demonstrations when the number of demonstrations is small.

% \section*{References}

\bibliographystyle{corlabbrvnat}
\bibliography{corl_2021}

\begin{thebibliography}{42}
\providecommand{\natexlab}[1]{#1}
\providecommand{\url}[1]{\texttt{#1}}
\expandafter\ifx\csname urlstyle\endcsname\relax
  \providecommand{\doi}[1]{doi: #1}\else
  \providecommand{\doi}{doi: \begingroup \urlstyle{rm}\Url}\fi

\bibitem[Kaelbling and Lozano-P{\'e}rez(2013)]{kaelbling2013integrated}
L.~P. Kaelbling and T.~Lozano-P{\'e}rez.
\newblock Integrated task and motion planning in belief space.
\newblock \emph{The International Journal of Robotics Research}, 32\penalty0
  (9-10):\penalty0 1194--1227, 2013.

\bibitem[Levine et~al.(2016)Levine, Finn, Darrell, and Abbeel]{levine2016end}
S.~Levine, C.~Finn, T.~Darrell, and P.~Abbeel.
\newblock End-to-end training of deep visuomotor policies.
\newblock \emph{The Journal of Machine Learning Research}, 17\penalty0
  (1):\penalty0 1334--1373, 2016.

\bibitem[Pathak et~al.(2017)Pathak, Agrawal, Efros, and
  Darrell]{pathak2017curiosity}
D.~Pathak, P.~Agrawal, A.~A. Efros, and T.~Darrell.
\newblock Curiosity-driven exploration by self-supervised prediction.
\newblock In \emph{International Conference on Machine Learning}, 2017.

\bibitem[Colas et~al.(2019)Colas, Oudeyer, Sigaud, Fournier, and
  Chetouani]{colas2019curious}
C.~Colas, P.-Y. Oudeyer, O.~Sigaud, P.~Fournier, and M.~Chetouani.
\newblock Curious: Intrinsically motivated modular multi-goal reinforcement
  learning.
\newblock In \emph{International Conference on Machine Learning}, pages
  1331--1340, 2019.

\bibitem[Dai et~al.(2021)Dai, Xu, Hofmann, and Williams]{DaiS-RSS-21}
S.~Dai, W.~Xu, A.~Hofmann, and B.~C. Williams.
\newblock {An Empowerment-based Solution to Robotic Manipulation Tasks with
  Sparse Rewards}.
\newblock In \emph{Proceedings of Robotics: Science and Systems}, Virtual, July
  2021.
\newblock \doi{10.15607/RSS.2021.XVII.001}.

\bibitem[Bacon et~al.(2017)Bacon, Harb, and Precup]{bacon2017option}
P.-L. Bacon, J.~Harb, and D.~Precup.
\newblock The option-critic architecture.
\newblock In \emph{Thirty-First AAAI Conference on Artificial Intelligence},
  2017.

\bibitem[Haarnoja et~al.(2018)Haarnoja, Hartikainen, Abbeel, and
  Levine]{haarnoja2018latent}
T.~Haarnoja, K.~Hartikainen, P.~Abbeel, and S.~Levine.
\newblock Latent space policies for hierarchical reinforcement learning.
\newblock In \emph{International Conference on Machine Learning}, pages
  1851--1860. PMLR, 2018.

\bibitem[Bengio et~al.(2009)Bengio, Louradour, Collobert, and
  Weston]{bengio2009curriculum}
Y.~Bengio, J.~Louradour, R.~Collobert, and J.~Weston.
\newblock Curriculum learning.
\newblock In \emph{Proceedings of the 26th annual international conference on
  machine learning}, pages 41--48. ACM, 2009.

\bibitem[Ivanovic et~al.(2019)Ivanovic, Harrison, Sharma, Chen, and
  Pavone]{ivanovic2019barc}
B.~Ivanovic, J.~Harrison, A.~Sharma, M.~Chen, and M.~Pavone.
\newblock Barc: Backward reachability curriculum for robotic reinforcement
  learning.
\newblock In \emph{2019 International Conference on Robotics and Automation
  (ICRA)}, pages 15--21. IEEE, 2019.

\bibitem[Ho and Ermon(2016)]{ho2016generative}
J.~Ho and S.~Ermon.
\newblock Generative adversarial imitation learning.
\newblock \emph{Advances in neural information processing systems},
  29:\penalty0 4565--4573, 2016.

\bibitem[Peng et~al.(2018)Peng, Kanazawa, Toyer, Abbeel, and
  Levine]{peng2018variational}
X.~B. Peng, A.~Kanazawa, S.~Toyer, P.~Abbeel, and S.~Levine.
\newblock Variational discriminator bottleneck: Improving imitation learning,
  inverse rl, and gans by constraining information flow.
\newblock In \emph{International Conference on Learning Representations}, 2018.

\bibitem[Salimans and Chen(2018)]{salimans2018learning}
T.~Salimans and R.~Chen.
\newblock Learning montezuma's revenge from a single demonstration.
\newblock \emph{arXiv preprint arXiv:1812.03381}, 2018.

\bibitem[Resnick et~al.(2018)Resnick, Raileanu, Kapoor, Peysakhovich, Cho, and
  Bruna]{resnick2018backplay}
C.~Resnick, R.~Raileanu, S.~Kapoor, A.~Peysakhovich, K.~Cho, and J.~Bruna.
\newblock Backplay:" man muss immer umkehren".
\newblock \emph{arXiv preprint arXiv:1807.06919}, 2018.

\bibitem[Graves et~al.(2017)Graves, Bellemare, Menick, Munos, and
  Kavukcuoglu]{graves2017automated}
A.~Graves, M.~G. Bellemare, J.~Menick, R.~Munos, and K.~Kavukcuoglu.
\newblock Automated curriculum learning for neural networks.
\newblock In \emph{Proceedings of the 34th International Conference on Machine
  Learning-Volume 70}, pages 1311--1320. JMLR. org, 2017.

\bibitem[Mirowski et~al.(2018)Mirowski, Grimes, Malinowski, Hermann, Anderson,
  Teplyashin, Simonyan, Zisserman, Hadsell, et~al.]{mirowski2018learning}
P.~Mirowski, M.~Grimes, M.~Malinowski, K.~M. Hermann, K.~Anderson,
  D.~Teplyashin, K.~Simonyan, A.~Zisserman, R.~Hadsell, et~al.
\newblock Learning to navigate in cities without a map.
\newblock In \emph{Advances in Neural Information Processing Systems}, pages
  2419--2430, 2018.

\bibitem[Florensa et~al.(2017)Florensa, Held, Wulfmeier, Zhang, and
  Abbeel]{florensa2017reverse}
C.~Florensa, D.~Held, M.~Wulfmeier, M.~Zhang, and P.~Abbeel.
\newblock Reverse curriculum generation for reinforcement learning.
\newblock In \emph{Conference on Robot Learning}, pages 482--495, 2017.

\bibitem[Wang et~al.(2019)Wang, Lehman, Clune, and Stanley]{wang2019paired}
R.~Wang, J.~Lehman, J.~Clune, and K.~O. Stanley.
\newblock Paired open-ended trailblazer (poet): Endlessly generating
  increasingly complex and diverse learning environments and their solutions.
\newblock \emph{arXiv preprint arXiv:1901.01753}, 2019.

\bibitem[Florensa et~al.(2018)Florensa, Held, Geng, and
  Abbeel]{florensa2018automatic}
C.~Florensa, D.~Held, X.~Geng, and P.~Abbeel.
\newblock Automatic goal generation for reinforcement learning agents.
\newblock In \emph{International conference on machine learning}, pages
  1515--1528, 2018.

\bibitem[Sukhbaatar et~al.(2018)Sukhbaatar, Lin, Kostrikov, Synnaeve, Szlam,
  and Fergus]{sukhbaatar2018intrinsic}
S.~Sukhbaatar, Z.~Lin, I.~Kostrikov, G.~Synnaeve, A.~Szlam, and R.~Fergus.
\newblock Intrinsic motivation and automatic curricula via asymmetric
  self-play.
\newblock In \emph{International Conference on Learning Representations}, 2018.
\newblock URL \url{https://openreview.net/forum?id=SkT5Yg-RZ}.

\bibitem[Pomerleau(1991)]{pomerleau1991efficient}
D.~A. Pomerleau.
\newblock Efficient training of artificial neural networks for autonomous
  navigation.
\newblock \emph{Neural computation}, 3\penalty0 (1):\penalty0 88--97, 1991.

\bibitem[Bain and Sammut(1995)]{bain1995framework}
M.~Bain and C.~Sammut.
\newblock A framework for behavioural cloning.
\newblock In \emph{Machine Intelligence 15}, pages 103--129, 1995.

\bibitem[Bansal et~al.(2019)Bansal, Krizhevsky, and Ogale]{Ogale-RSS-19}
M.~Bansal, A.~Krizhevsky, and A.~Ogale.
\newblock Chauffeurnet: Learning to drive by imitating the best and
  synthesizing the worst.
\newblock In \emph{Proceedings of Robotics: Science and Systems},
  FreiburgimBreisgau, Germany, June 2019.
\newblock \doi{10.15607/RSS.2019.XV.031}.

\bibitem[Wang et~al.(2019)Wang, Devin, Cai, Yu, and Darrell]{wang2019deep}
D.~Wang, C.~Devin, Q.-Z. Cai, F.~Yu, and T.~Darrell.
\newblock Deep object-centric policies for autonomous driving.
\newblock In \emph{2019 International Conference on Robotics and Automation
  (ICRA)}, pages 8853--8859. IEEE, 2019.

\bibitem[Rahmatizadeh et~al.(2018)Rahmatizadeh, Abolghasemi, B{\"o}l{\"o}ni,
  and Levine]{rahmatizadeh2018vision}
R.~Rahmatizadeh, P.~Abolghasemi, L.~B{\"o}l{\"o}ni, and S.~Levine.
\newblock Vision-based multi-task manipulation for inexpensive robots using
  end-to-end learning from demonstration.
\newblock In \emph{2018 IEEE international conference on robotics and
  automation (ICRA)}, pages 3758--3765. IEEE, 2018.

\bibitem[Ross et~al.(2011)Ross, Gordon, and Bagnell]{ross2011reduction}
S.~Ross, G.~Gordon, and D.~Bagnell.
\newblock A reduction of imitation learning and structured prediction to
  no-regret online learning.
\newblock In \emph{Proceedings of the fourteenth international conference on
  artificial intelligence and statistics}, pages 627--635, 2011.

\bibitem[Ng et~al.(2000)Ng, Russell, et~al.]{ng2000algorithms}
A.~Y. Ng, S.~J. Russell, et~al.
\newblock Algorithms for inverse reinforcement learning.
\newblock In \emph{Icml}, volume~1, page~2, 2000.

\bibitem[Finn et~al.(2016)Finn, Levine, and Abbeel]{finn2016guided}
C.~Finn, S.~Levine, and P.~Abbeel.
\newblock Guided cost learning: Deep inverse optimal control via policy
  optimization.
\newblock In \emph{International conference on machine learning}, pages 49--58,
  2016.

\bibitem[Fu et~al.(2017)Fu, Luo, and Levine]{fu2017learning}
J.~Fu, K.~Luo, and S.~Levine.
\newblock Learning robust rewards with adversarial inverse reinforcement
  learning.
\newblock \emph{arXiv preprint arXiv:1710.11248}, 2017.

\bibitem[Arjovsky et~al.(2017)Arjovsky, Chintala, and
  Bottou]{arjovsky2017wasserstein}
M.~Arjovsky, S.~Chintala, and L.~Bottou.
\newblock Wasserstein generative adversarial networks.
\newblock In \emph{International conference on machine learning}, pages
  214--223. PMLR, 2017.

\bibitem[Vecerik et~al.(2017)Vecerik, Hester, Scholz, Wang, Pietquin, Piot,
  Heess, Roth{\"o}rl, Lampe, and Riedmiller]{vecerik2017leveraging}
M.~Vecerik, T.~Hester, J.~Scholz, F.~Wang, O.~Pietquin, B.~Piot, N.~Heess,
  T.~Roth{\"o}rl, T.~Lampe, and M.~Riedmiller.
\newblock Leveraging demonstrations for deep reinforcement learning on robotics
  problems with sparse rewards.
\newblock \emph{arXiv preprint arXiv:1707.08817}, 2017.

\bibitem[Kang et~al.(2018)Kang, Jie, and Feng]{kang2018policy}
B.~Kang, Z.~Jie, and J.~Feng.
\newblock Policy optimization with demonstrations.
\newblock In \emph{International Conference on Machine Learning}, pages
  2469--2478. PMLR, 2018.

\bibitem[Nair et~al.(2018)Nair, McGrew, Andrychowicz, Zaremba, and
  Abbeel]{nair2018overcoming}
A.~Nair, B.~McGrew, M.~Andrychowicz, W.~Zaremba, and P.~Abbeel.
\newblock Overcoming exploration in reinforcement learning with demonstrations.
\newblock In \emph{2018 IEEE International Conference on Robotics and
  Automation (ICRA)}, pages 6292--6299. IEEE, 2018.

\bibitem[Andrychowicz et~al.(2017)Andrychowicz, Wolski, Ray, Schneider, Fong,
  Welinder, McGrew, Tobin, Abbeel, and Zaremba]{andrychowicz2017hindsight}
M.~Andrychowicz, F.~Wolski, A.~Ray, J.~Schneider, R.~Fong, P.~Welinder,
  B.~McGrew, J.~Tobin, O.~P. Abbeel, and W.~Zaremba.
\newblock Hindsight experience replay.
\newblock In \emph{Advances in Neural Information Processing Systems}, pages
  5048--5058, 2017.

\bibitem[Eysenbach et~al.(2018)Eysenbach, Gu, Ibarz, and
  Levine]{eysenbach2018leave}
B.~Eysenbach, S.~Gu, J.~Ibarz, and S.~Levine.
\newblock Leave no trace: Learning to reset for safe and autonomous
  reinforcement learning.
\newblock In \emph{International Conference on Learning Representations}, 2018.
\newblock URL \url{https://openreview.net/forum?id=S1vuO-bCW}.

\bibitem[Xu et~al.(2020)Xu, Verma, Finn, and Levine]{xu2020continual}
K.~Xu, S.~Verma, C.~Finn, and S.~Levine.
\newblock Continual learning of control primitives : Skill discovery via
  reset-games.
\newblock In H.~Larochelle, M.~Ranzato, R.~Hadsell, M.~F. Balcan, and H.~Lin,
  editors, \emph{Advances in Neural Information Processing Systems}, volume~33,
  pages 4999--5010, 2020.

\bibitem[Turchetta et~al.(2020)Turchetta, Kolobov, Shah, Krause, and
  Agarwal]{turchetta2020safe}
M.~Turchetta, A.~Kolobov, S.~Shah, A.~Krause, and A.~Agarwal.
\newblock Safe reinforcement learning via curriculum induction.
\newblock In H.~Larochelle, M.~Ranzato, R.~Hadsell, M.~F. Balcan, and H.~Lin,
  editors, \emph{Advances in Neural Information Processing Systems}, volume~33,
  pages 12151--12162, 2020.

\bibitem[Sutton and Barto(2018)]{sutton2018reinforcement}
R.~S. Sutton and A.~G. Barto.
\newblock \emph{Reinforcement learning: An introduction}.
\newblock MIT press, 2018.

\bibitem[Schulman et~al.(2017)Schulman, Wolski, Dhariwal, Radford, and
  Klimov]{schulman2017proximal}
J.~Schulman, F.~Wolski, P.~Dhariwal, A.~Radford, and O.~Klimov.
\newblock Proximal policy optimization algorithms.
\newblock \emph{arXiv preprint arXiv:1707.06347}, 2017.

\bibitem[Lillicrap et~al.(2015)Lillicrap, Hunt, Pritzel, Heess, Erez, Tassa,
  Silver, and Wierstra]{lillicrap2015continuous}
T.~P. Lillicrap, J.~J. Hunt, A.~Pritzel, N.~Heess, T.~Erez, Y.~Tassa,
  D.~Silver, and D.~Wierstra.
\newblock Continuous control with deep reinforcement learning.
\newblock \emph{arXiv preprint arXiv:1509.02971}, 2015.

\bibitem[Brockman et~al.(2016)Brockman, Cheung, Pettersson, Schneider,
  Schulman, Tang, and Zaremba]{1606.01540}
G.~Brockman, V.~Cheung, L.~Pettersson, J.~Schneider, J.~Schulman, J.~Tang, and
  W.~Zaremba.
\newblock Openai gym, 2016.

\bibitem[Lanier et~al.(2019)Lanier, McAleer, and Baldi]{lanier2019curiosity}
J.~B. Lanier, S.~McAleer, and P.~Baldi.
\newblock Curiosity-driven multi-criteria hindsight experience replay.
\newblock \emph{arXiv preprint arXiv:1906.03710}, 2019.

\bibitem[Dhariwal et~al.(2017)Dhariwal, Hesse, Klimov, Nichol, Plappert,
  Radford, Schulman, Sidor, Wu, and Zhokhov]{baselines}
P.~Dhariwal, C.~Hesse, O.~Klimov, A.~Nichol, M.~Plappert, A.~Radford,
  J.~Schulman, S.~Sidor, Y.~Wu, and P.~Zhokhov.
\newblock Openai baselines.
\newblock \url{https://github.com/openai/baselines}, 2017.

\end{thebibliography}

%%%%%%%%%%%%%%%%%%%%%%%%%%%%%%%%%%%%%%%%%%%%%%%%%%%%%%%%%%%%

%%%%%%%%%%%%%%%%%%%%%%%%%%%%%%%%%%%%%%%%%%%%%%%%%%%%%%%%%%%%
\clearpage

\appendix

\section{Implementation Details}  \label{sec:details}

All experiments shown in this paper are conducted on a 10-core Intel i7 3.0 GHz desktop with 64 GB RAM and one GeForce GTX 1080 GPU. In our implementation of the PPO algorithm~\cite{schulman2017proximal}, we use a three hidden-layer fully-connect neural network with (128, 64, 32) units in each layer for both the policy network and the value network, and set $\gamma = 0.99$ and $\lambda = 0.95$. We noticed that PPO training can be unstable due to the frequent curriculum switches especially in the block stacking environment, and found that in order to prevent collapsing during training, it is very helpful to use a small importance ratio clipping parameter in PPO (denoted as $\epsilon$ in \cite{schulman2017proximal}) together with an optimizer with small learning rates and gradient clipping. In pick-and-place tasks, we set $\epsilon = 0.2$ and use the Adam optimizer with a learning rate of 2e-4 without gradient clipping. In stacking tasks, we set $\epsilon = 0.05$ and use the Adam optimizer with a learning rate of 1e-4 and gradient clipping-by-norm with a clipping factor of 0.05. In our DDPG~\cite{lillicrap2015continuous} implementation, the learning rate is set to 2e-4, and the same policy network is used as in the PPO implementation. The target update period in DDPG is set to 5.

In the ACED algorithm, we use $\phi = 0.9$ as the curriculum switching threshold in pick-and-place tasks, and $\phi = 0.85$ in block stacking tasks. The average return checking period is set to $t = 120$, and the number of episodes used to compute the average return is set to $n=3$. 60 parallel rollout workers are used in both tasks. In pick-and-place tasks, the threshold for the object's distance to the goal to assign reward $r=1$ is 0.05, and in stacking tasks the threshold is set to 0.04. The maximum number of steps in an episode is set to 50 in pick-and-place tasks, and 100 in block stacking tasks. In BC, an Adam optimizer with the learning rate of 2e-4 is used, and the loss function is negative log likelihood.

In the reverse curriculum implementation, we use 1000 random start states and a time horizon of 5 time steps for the Brownian motion. 200 old sampled start states are appended to the new start states at each training step. In the our implementation of the Montezuma's Revenge method, we randomly select one demonstration trajectory and set the curriculum switching threshold also to  $\phi = 0.85$. Both the reverse curriculum method and the Montezuma's Revenge method are implemented with the same PPO algorithm as used in the ACED implementation.

\section{Additional Results} \label{sec:learning_curves}

Due to limited space, additional experimental results are presented here in the Appendix. Since each experiment is terminated after convergence, the lengths of the learning curves may vary.

\subsection{Learning Curves}

\begin{figure}
 \begin{center}
 \includegraphics[width=0.8\linewidth]{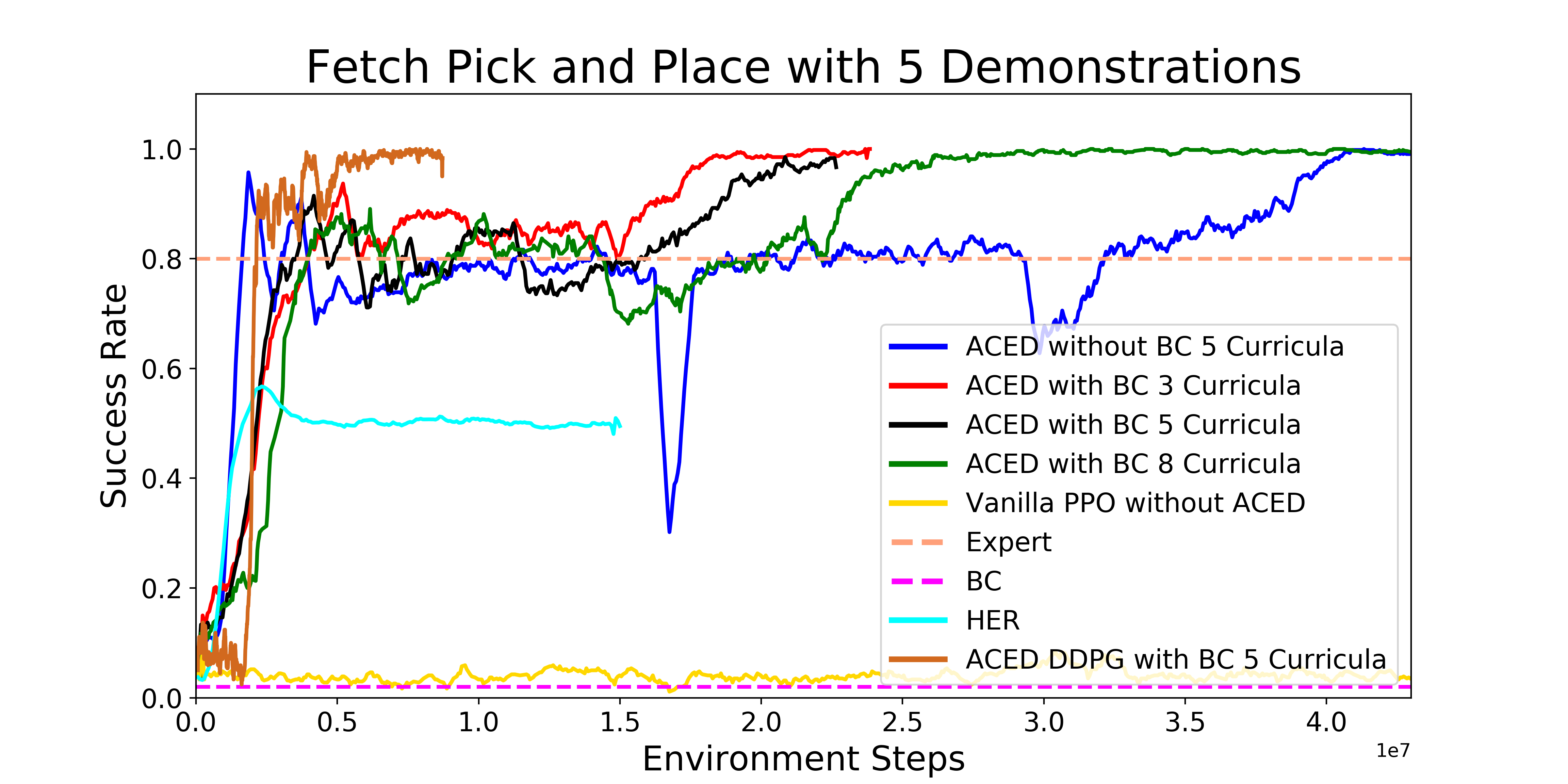}
 \caption{Learning curves of different algorithms in the pick-and-place environment with 5 demonstration trajectories. The horizontal axis represents the number of environment steps during training and the vertical axis represents the success rate. Expert and BC success rates are represented by dash lines because they didn't have training processes and their success rates remain constant.}
 \label{fig:learning_curves}
 \end{center}
\end{figure}

In order to show the learning progress and curriculum switches when using ACED, we use the pick-and-place task with 5 demonstration trajectories as an example to compare the learning curves of different algorithms, as shown in Figure~\ref{fig:learning_curves}. For each algorithm, we select one run whose convergence environment step is close to the mean for all 10 runs instead of directly using the mean in order to clearly show the learning progress and the curriculum switches. From Figure~\ref{fig:learning_curves} we can see that for all ACED runs with PPO, the first few curricula are usually much easier than the last few and the majority of training time is spent on training the last few curricula. Without BC, the performance drop during curriculum switches is more obvious. If we compare the performance of ACED with DDPG and ACED with PPO, we can observe that ACED achieves a much higher sample efficiency with the off-policy DDPG. 

All ACED runs are able to converge to almost 100\% success rate, whereas vanilla PPO without ACED is not able to achieve a success rate higher than 10\% during training. In addition to the comparison with vanilla PPO, we also compare ACED with Hindsight Experience Replay (HER)~\cite{andrychowicz2017hindsight} in the block pick-and-place task. We use the OpenAI Baseline~\cite{baselines} implementation of HER with 2 MPI processes with 30 parallel environments each to make sure it is equivalent to the 60 parallel environments in other experiments. Other parameters for HER are set to default. However, all 10 runs with HER are only able to achieve a success rate of about 50\%, and we show one representative learning curve in Figure~\ref{fig:learning_curves}. This is because in the Gym FetchPickAndPlace-V1 task, half of the goals are sampled from on the table and half are sampled in the air, thus agents that only learned to push can still reach the goals close to the tabletop and receive a success rate of about 50\%, but only agents that actually learned to pick and place will reach a success rate of 100\%.

\subsection{Comparison with BC + RL without ACED}

In order to further demonstrate the role of curriculum learning in ACED, this section compares the performance of ACED with an algorithm that only uses BC pre-trained policies to initialize the RL agent but doesn't use curriculum learning during RL training. We refer to the RL algorithm that uses BC to pre-train the policy but doesn't use ACED as ``BC + RL''. The same PPO algorithm and BC pre-trained policy initializations as in the ACED experiments are used in all experiments presented in this section. We summarizes the performance of both ACED with BC and BC + RL in Table~\ref{table:BC_RL} for convenient comparison, but the ACED with BC data in Table~\ref{table:BC_RL} are the same as the ones presented in Figure~\ref{fig:BarChart} and Table~\ref{table:SuccessRate}. As shown in Table~\ref{table:BC_RL}, BC + RL only works better than ACED with BC when $|\mathcal{T}|=100$, whereas its performance in terms of both convergence speed and success rate is worse than that of ACED with BC when $|\mathcal{T}|=50$ and $|\mathcal{T}|=20$. When $|\mathcal{T}|=5$ or $|\mathcal{T}|=1$, BC + RL is not able to learn pick-and-place and none of the runs converged to a success rate of 100\%. Recorded videos show that when $|\mathcal{T}|=5$ and $|\mathcal{T}|=1$, BC + RL can only learn to push the block to goal poses that are on the tabletop, but failed to learn pick-and-place when the goal pose is in the air. Since the goal pose has a 50\% probability of being in the air in the pick-and-place environment, all the runs have converged to a success rate of around 50\% during training.

We also evaluated BC + RL in the block stacking environment, but results show that none of the runs with $|\mathcal{T}|=100$ or $|\mathcal{T}|=20$ can converge to a success rate of 100\%. In fact, the training curves remain zero throughout the entire training progress for all runs with BC + RL. The comparison between ACED with BC and BC + RL shows that ACED is especially helpful in scenarios where the target task is complicated or the number of demonstrations is small.

\begin{table}
\caption{Pick-and-Place Comparison with BC + RL}
\label{table:BC_RL}
\centering
\begin{threeparttable}
\begin{tabular}{L{1.4cm}|L{1.9cm} L{1.6cm}|L{1.5cm} L{1.4cm} L{1.4cm} L{1.4cm} L{1.4cm}}
\toprule

 \multicolumn{3}{c|}{Algorithm\tnote{1}}  & $|\mathcal{T}|=100$  & $|\mathcal{T}|=50$  & $|\mathcal{T}|=20$  & $|\mathcal{T}|=5$\tnote{2}  & $|\mathcal{T}|=1$\tnote{2}   \\ 

\hline

\multirow{8}{1.4cm}{\centering ACED with BC} & \multirow{4}{1.9cm}{\centering Convergence Env Steps (Million)} & $C_{max}=8$ & 2.78 & 3.40  & 8.40  & 24.00 & 41.21 \\
& & $C_{max}=5$ & 2.32 & 3.55  & 5.97  & 16.50 & 38.85 \\
& & $C_{max}=3$ & 4.15 & 6.53  & 7.88  & 17.67 & 25.56 \\

\cline{3-8}

& & Average\tnote{3} & 3.08 & 4.49  & 7.41  & 19.39 & 35.21 \\ \cline{2-8}

& \multirow{4}{1.9cm}{\centering Success Rate} & $C_{max}=8$ & 99\% & 100\%  & 99\%  & 97\% & 96\% \\
& & $C_{max}=5$ & 96\% & 99\%  & 99\%  & 100\% & 95\% \\
& & $C_{max}=3$ & 100\% & 99\%  & 100\%  & 100\% & 99\% \\

\cline{3-8}

& & Average\tnote{3} & 98.3\% & 99.3\%  & 99.3\%  & 99\% & 96.7\% \\ \hline

\multirow{2}{1.4cm}{\centering BC + RL} & \multicolumn{2}{c|}{Convergence Env Steps} & 2.47 & 14.66 & 13.58 & (67.67) & (61.73) \\
& \multicolumn{2}{c|}{Success Rate} & 100\% & 97\% & 99\% & (37\%) & (53\%) \\

% ACED without BC & $C_{max}=5$ & 100\% & 95\%  & 100\%  & 93\% & 97\% \\
% \hline
% \multicolumn{2}{c|}{BC Policy\tnote{3}} & 60\% & 54\% & 24\% & 2\% & 4\% \\
% \multicolumn{2}{c|}{Expert Demonstrations} & 92\% & 98\%  & 95\%  & 80\% & 100\% \\

\bottomrule
\end{tabular}
\begin{tablenotes}
\footnotesize
 \item[1] For each set of experiment except for BC + RL with $|\mathcal{T}|=5$ and $|\mathcal{T}|=1$, we have 10 runs with different random seeds and the entries in the table are averaged from all runs. For BC + RL with $|\mathcal{T}|=5$ and $|\mathcal{T}|=1$, we only conducted 3 runs each due to their long training time. For each run, we rollout 10 trajectories with the policy at convergence, and we compute the success rate by taking the average of all rollout trajectories for all runs. 
 \item[2] The entries for BC + RL with $|\mathcal{T}|=5$ and $|\mathcal{T}|=1$ are in brackets because none of these experiments have actually converged to a success rate of 100\% during training. They instead converged to around 50\% because they have only learned to push the block to the goal when the goal pose is on the tabletop, but they failed to learn how to pick up the block and lift them to the goal poses that are in the air.
 \item[3] The average success rate for $C_{max}=8$, $C_{max}=5$ and $C_{max}=3$.
\end{tablenotes}
\end{threeparttable}
\end{table}

\end{document}